# Using 3D Hahn Moments as A Computational Representation of ATS Drugs Molecular Structure

Satrya Fajri Pratama[1], Azah Kamilah Muda[1,✉], Yun-Huoy Choo[1], Ramon Carbó-Dorca[2], Ajith Abraham[1,3]

[1]*Computational Intelligence and Technologies (CIT) Research Group,*
*Center of Advanced Computing and Technologies,*
*Faculty of Information and Communication Technology,*
*Universiti Teknikal Malaysia Melaka*
*Hang Tuah Jaya, 76100 Durian Tunggal, Melaka, Malaysia*
E-mail: `satrya@student.utem.edu.my, {azah, huoy}@utem.edu.my, ramoncarbodorca@gmail.com`
`ajith.abraham@ieee.org`

[2]*Institut de Química Computacional i Catàlisi, Universitat de Girona*
*17071 Girona, Catalonia, Spain*

[3]*Machine Intelligence Research Labs (MIR Labs)*
*Scientific Network for Innovation and Research Excellence, Auburn, WA, USA*

*Abstract*— The campaign against drug abuse is fought by all countries, most notably on ATS drugs. The technical limitations of the current test kits to detect new brand of ATS drugs present a challenge to law enforcement authorities and forensic laboratories. Meanwhile, new molecular imaging devices which allowed mankind to characterize the physical 3D molecular structure have been recently introduced, and it can be used to remedy the limitations of existing drug test kits. Thus, a new type of 3D molecular structure representation technique should be developed to cater the 3D molecular structure acquired physically using these molecular imaging devices. One of the applications of image processing methods to represent a 3D image is 3D moments, and this study formulates a new 3D moments technique, namely 3D Hahn moments, to represent the 3D molecular structure of ATS drugs. The performance of the proposed technique was analysed using drug chemical structures obtained from UNODC for the ATS drugs, while non-ATS drugs are obtained randomly from ChemSpider database. The evaluation shows the technique is qualified to be further explored in the future works to be fully compatible with ATS drug identification domain.

*Keywords*— 3D moments; ATS drugs; drugs identification; Hahn moments; molecular similarity; molecular structure representation.

## I. INTRODUCTION

Every country in this world is continuously threaten by the destructive potential of ATS drugs abuse. One of the main reasons the ATS drugs abuse is that it is mostly synthetic in origin [1]. Therefore, provided with sufficient technical knowledge, a chemist can derive a novel and unregulated structure of ATS, often referred as designer drugs, and manufacture it in the clandestine laboratories [2]. These innovative designs proven to be a challenge for law enforcement agencies and introduce a new layer of difficulties in detecting and identifying using existing laboratory techniques, which sometimes lead to both false positive and negative results, and thus, incorrectly outline the outcome of criminal justice court [3]–[6]. Due to these limitations, it is preferable to perform the identification by relying on computational intelligence methods [7], and taking advantage of the shape of molecular structures.

Despite the apparent need of reliable methods to detect and identify ATS drugs in the field of forensic toxicology, the use of computer-assisted methods is only starting to be incorporated recently [8]–[14]. The use of computer science in the field of chemistry breeds a new domain namely cheminformatics. Cheminformatics researchers use molecular similarity search to seek structurally similar compounds, based on a principle which states that structurally similar compounds are more likely to exhibit comparable properties [15]–[19]. The success of molecular similarity search depends on the molecular structures representation employed, also known as the molecular descriptors [20], [21]. There are various categories of molecular descriptors, with the most commonly used molecular descriptors are 2D or topological molecular descriptors and 3D or geometrical molecular descriptors.

Since a molecular structure representation involves the knowledge of the relative positions of the atoms in 3D space, geometrical descriptors usually provide more information and discrimination power for similar molecular structures and molecule conformations than topological descriptors [22]–[25]. This study believes that molecular similarity can also be used to detect the similarity between unknown compound and reference compounds, and thus it can be used to detect new brand of ATS drugs to the known samples of ATS drugs based on their molecular structure [8], [9], [12].

Recent advances in scientific domains have enabled the characterization of the physical molecular structure through molecular microscopy, such as by using transmission electron microscopy (TEM), scanning tunnelling microscopy (STM), and more recently, non-contact atomic force microscope (nc-AFM) [26]. New discovery of tuning-fork-based nc-AFM provides a novel method capable of non-destructive sub-nanometer spatial resolution [27]–[31]. Single-molecule images obtained with this technique are reminiscent of wire-frame chemical structures and even allow differences in chemical bond-order to be identified [29]. The high-resolution images obtained reveals that the 3D model which has been used for decades to depict molecular structure is virtually identical to the physical molecules.

Invariance with respect to labelling, numbering of the molecule atoms, and molecule translation and rotation is a required property of a molecular descriptor. Furthermore, it also must have a clear algorithmically quantifiable definition, and the values must be in an appropriate numerical range for the molecule set where it is applicable to [32], [33]. Since a molecular descriptor is independent of the characteristics of the molecular representation, it is possible to consider the molecular shape as an image, and thus apply image processing to represent the shape of the molecular structure.

One of the applications of image processing methods to represent 2D and 3D image is moment invariants (MI), which can easily achieve these invariance properties. MI is a special case of moments. Moments are scalar quantities used to characterize a function and to capture its crucial features, or from the mathematical point of view, moments are projections of a function onto a polynomial basis [34]. The first application of MI as molecular descriptors is 3D Zernike descriptors, although it was introduced to represent the molecular surface of protein structure [35].

There are several advantages of MI-based molecular shape representation compared to conventional representations. First, MI-based descriptors allow for fast retrieval and comparison of molecular structures. Second, due to its rotation and translation invariance properties, molecular structures need not be aligned for comparison. Lastly, the resolution of the description of molecular structures can be easily and naturally adjusted by changing the order of shape descriptors [35], [36].

However, deriving MI from a moments technique is rather difficult, since it must satisfy the invariance requirements where an object representation must be invariant when said object is underwent translation, scale, and rotation transformations [37]. Therefore, this study only aims to propose a moments-based instead of MI-based molecular representation technique. This is because the development of MI-based molecular representation technique will be conducted extensively in the future works, and this study will be used as the basis of the MI-based technique, should this study exhibit satisfactory results.

Before the proposed technique is described and the justification of the results produced is presented, the existing moments-based techniques, which are also going to be used as the comparison for the proposed technique, must be discussed first. Thus, the remainder of the paper is organized as follows. The ensuing section will provide the material and method, while the results and conclusion are discussed in Sections 3 and 4, respectively.

## II. THE MATERIAL AND METHOD

In the following subsections, an overview of existing 3D moments is provided, the proposed technique is introduced, and the experimental setup describing the data source collection and experimental design are presented, respectively.

### A. Existing 3D Moments Techniques

Representations of formal molecular shapes and surfaces provide more detailed and more chemically relevant information than simple molecular graphs and stereochemical bond structures and give a more faithful description of actual molecular recognition and interaction processes, which lead to the quantitative shape-activity relations (QShAR) domain [38]. Shape is an important visual feature and it is one of the basic features used to describe image content [39], and hence, searching for an image by using the shape features gives challenges for many researches, since extracting the features that represent and describe the shape is an arduous task [40], which also hold true in the QShAR domain [38].

In the QShAR domain itself, various molecular shape representation techniques have been proposed [38], [41]. Meanwhile, in pattern recognition problem, there are many shape representations or description techniques have been explored to extract the features from the object. A decent shape descriptor should be able to find perceptually similar shapes, akin to human beings in comparing the object shapes. One of the most commonly used shape descriptors is moments, because it can satisfy this requirement.

Moments can be used to generate a set of numbers that uniquely represent the global characteristic of an image, and has been used in diverse fields ranging from mechanics and statistics to pattern recognition and image understanding [42]. The use of moments to calculate image features in image analysis and pattern recognition was inspired by [43] and [44].

A 2D image is considered as piece-wise continuous real function $f(x, y)$ of two variables defined on a compact support $D \subset R \times R$ and having a finite nonzero integral, which by extension is also applicable to 3D images. Moments of a 3D image $f(x, y, z)$ are usually denoted by

$$M_{pqr} = \iiint_D p_{pqr}(x,y,z) f(x,y,z) \, dx \, dy \, dz \quad (1)$$

where $p, q, r$ are non-negative integers, $s = p + q + r$ is called the order of the moment, and $p_{pqr}(x, y, z)$ are polynomial basis functions defined on $D$. Based on the polynomial basis

$p_{pqr}(x,y,z)$ used, there are various systems of moments which can be recognized. Additionally, if the polynomial basis $p_{pqr}(x,y,z)$ is orthogonal, specifically if its elements satisfy the condition of (weighted) orthogonality

$$\iiint_\Omega \frac{w(x,y,z)p_{pqr}(x,y,z)p_{mno}(x,y,z)}{p_{ijk}(x,y,z)} \, dx \, dy \, dz = 0 \quad (2)$$

for any indices $p \neq m \neq i$ or $q \neq n \neq j$ or $r \neq o \neq k$ and $\Omega$ is the area of orthogonality, the moments are categorized as orthogonal moments [45]–[48]. However, the weight function $w(x,y)$ in some orthogonal moments is not required. Some of the existing and well-known moments, including orthogonal moments, are discussed in the following subsections.

1) *Geometric Moments:* The simplest choice of the standard power basis as polynomial basis $p_{pq}(x,y) = x^p y^q$ leads to geometric moments, first introduced by [43], which is defined as

$$m_{pq} = \int_{-\infty}^{\infty} \int_{-\infty}^{\infty} x^p y^q f(x,y) \, dx \, dy \quad (3)$$

where $p, q = 0,1,2 \ldots$. The formula can also be generalized to 3D geometric moments [49]

$$m_{pqr} = \int_{-\infty}^{\infty} \int_{-\infty}^{\infty} \int_{-\infty}^{\infty} x^p y^q z^r f(x,y,z) \, dx \, dy \, dz \quad (4)$$

A digital 3D image of size $N \times N \times N$ is an array of voxels (volume pixels), therefore the triple integral in (4) must be replaced by triple summation. The most common way is to employ the zero-order method of numeric integration. And thus, (4) takes the following discrete form

$$\hat{m}_{pqr} = \sum_{i=1}^{N} \sum_{j=1}^{N} \sum_{k=1}^{N} i^p j^q k^r f_{ijk} \quad (5)$$

where $i, j, k$ are coordinates of the voxels and $f_{ijk}$ is the gray-level of the voxel $i,j,k$. It should be noted that $\hat{m}_{pqr}$ is just an approximation of $m_{pqr}$, thus introduced the approximation errors. Despite there are various methods to minimize the approximation errors when calculating 3D geometric moments, the study conducted by [50] had shown that the precise estimation is more preferred than the exact computation. The precise computation was proposed by [51], which is the 3D extension of a formula proposed by [52] generalized from [53], where the authors integrate the monomials $x^p y^q z^r$ precisely by the Newton–Leibnitz formula on each pixel

$$\dot{m}_{pqr} = \sum_{i=1}^{N} \sum_{j=1}^{N} \sum_{k=1}^{N} f_{ijk} \iiint_{A_{ijk}} x^p y^q z^r \, dx \, dy \, dz$$
$$= \sum_{i=1}^{N} \sum_{j=1}^{N} \sum_{k=1}^{N} U_p(i) U_q(j) U_r(k) f_{ijk} \quad (6)$$

$$U_s(a) = \frac{(a+0.5)^{s+1} - (a-0.5)^{s+1}}{s+1} \quad (7)$$

where $A_{ijk}$ denotes the area of the voxel $i,j,k$. Eq. (6) is still a zero-order approximation of moments of the original image, but only the image is approximated, the monomials are integrated exactly.

2) *Complex Moments:* Complex moments were originally proposed by [54] and later extended by [55] due to very little researches devoted to the independence of the invariants. The independence of the features is a fundamental issue in all the pattern recognition problems, especially in the case of a high-dimensional feature space. Recently, [56] proposed a 3D complex moments as the extension to the original complex moments. The 3D complex moments can be defined as projections on the corresponding spherical harmonics times $\varrho^s$

$$c_{sl}^m$$
$$= \int_0^{2\pi} \int_0^{\pi} \int_0^{\infty} \varrho^{s+2} Y_l^m(\theta, \varphi) \sin\theta \, f(\varrho, \theta, \varphi) \, d\varrho \, d\theta \, d\varphi \quad (8)$$
$$s = 0,1,\ldots$$
$$l = \begin{cases} 0,2,4,\ldots,s-2,s & s \text{ is even} \\ 1,3,5,\ldots,s-2,s & s \text{ is odd} \end{cases}$$
$$m = -l, -l+1, \ldots, l$$

where $\varrho, \theta, \varphi$ is the spherical coordinates, $s$ is the order of the moment, $l$ is called latitudinal repetition, $m$ is called longitudinal repetition, $\varrho^2 \sin\theta$ is the Jacobian of the transformation of Cartesian to spherical coordinates $\varrho, \theta, \varphi$, and $Y_l^m(\theta, \varphi)$ is the spherical harmonics given as

$$Y_l^m(\theta, \varphi) = \sqrt{\frac{2l+1}{4\pi} \frac{(l-m)!}{(l+m)!}} P_l^m(\cos\theta) \, e^{i\varphi} \quad (9)$$

where $P_l^m$ is an associated Legendre function defined as

$$P_l^m(a) = (-1)^m (1-a^2)^{\frac{m}{2}} \left(\frac{d}{da}\right)^m L_l(a) \quad (10)$$

and $L_l(a)$ is a Legendre polynomial defined as

$$L_s(a) = \sum_{k=0}^{s} c_{k,s} a^k = \frac{(-1)^s}{2^s s!} \left(\frac{d}{da}\right)^s [(1-a^2)^s] \quad (11)$$

Since $Y_l^{-m} = (-1)^m \overline{Y_l^m}$, it can be derived that $c_{sl}^{-m} = (-1)^m \overline{c_{sl}^m}$.

3) *Legendre Moments:* Legendre moments were proposed by [57] because Legendre moments can be used to represent an image in Cartesian domain, with a near zero value of information redundancy [58]. Legendre moments of order $p + q$ is defined as [59]

$$\lambda_{pq} = \frac{(2p+1)(2q+1)}{4} \int_{-1}^{1} \int_{-1}^{1} L_p(x) L_q(y) f(x,y) \, dx \, dy \quad (12)$$

where $p, q = 0,1, \ldots$. The $s$th-order Legendre polynomials are defined in (11), which can also be written as

$$L_s(a) = \sum_{k=0}^{\left\lfloor \frac{s}{2} \right\rfloor} (-1)^k \frac{(2s-2k)!}{2^s k! (s-k)! (s-2k)!} a^{s-2k} \quad (13)$$

The set of Legendre polynomials $L_s(a)$ forms a complete orthogonal basis set on the interval $[-1,1]$

$$\int_{-1}^{1} L_p(a)L_q(a)\,da = \frac{2}{2p+1}\delta_{pq} \quad (14)$$

where $\delta_{pq}$ is the Kronecker delta.

3D Legendre moments was proposed by [60] as the extension of 2D Legendre moments, which was derived directly from Legendre polynomials. Like 2D Legendre, the values must be scaled in the region of $-1 \leq x, y, z \leq 1$. The equations of 3D Legendre moments are defined as

$$\lambda_{pqr} = \frac{(2p+1)(2q+1)(2r+1)}{8} \times \int_{-1}^{1}\int_{-1}^{1}\int_{-1}^{1} L_P(x)L_q(y)L_r(z)f(x,y,z)\,dx\,dy\,dz \quad (15)$$

*4) Zernike Moments:* The set of orthogonal Zernike moments was first introduced for image analysis by [57]. Although it is computationally complex, Zernike moments had been proven to be superior in terms of their feature representation capability, image reconstruction capability, and low noise sensitivity [61]. Besides that, the orthogonal property also enables the separation of the individual contributions of each order moment to the reconstruction process. The 3D Zernike moments was first proposed to 3D by [62]. The 3D Zernike moments can be determined by using the complex conjugate of 3D Zernike polynomials defined as

$$\Omega_{nl}^m(R) = \int_0^1 \int_0^{2\pi} \int_0^\pi \overline{Z_{nl}^m(R)} f(R)\varrho^2 \sin\theta\,d\varrho\,d\theta\,d\varphi \quad (16)$$

The 3D unit-ball Zernike polynomials in spherical coordinates is defined as

$$Z_{nl}^m(R) = R_{nl}(\varrho)Y_l^m(\theta,\varphi) \quad (17)$$

where $0 \leq l \leq n$, $-l \leq m \leq l$, $n-l$ is an even non-negative integer number, and $R = (\varrho, \theta, \varphi)^T$ is the spherical coordinates. $R_{nl}(\varrho)$ is the real-valued radial functions, and $Y_l^m(\theta, \varphi)$ is the spherical harmonics given in (9). Spherical harmonics are orthonormal on the surface of the unit sphere per the relation

$$\int_0^\pi \int_0^{2\pi} Y_l^m(\theta,\varphi)\overline{Y_{l'}^{m'}(\theta,\varphi)} \sin\theta\,d\theta\,d\varphi = \delta_{ll'}\delta^{mm'} \quad (18)$$

$R_{nl}(\varrho)$ are radial functions constructed by [62] to redefine the original Zernike polynomials in Cartesian coordinates as

$$Z_{nl}^m(X) = \sum_{v=0}^{k} q_{kl}^v \|X\|^{2v} e_l^m(X) \quad (19)$$

where $2k = n - l$, $0 \leq v \leq k$, and $X$ denotes the vector $X = (x, y, z)^T$. Here, $e_l^m$ are the harmonic polynomials defined as

$$e_l^m(X) = \varrho^l Y_l^m(\theta,\varphi)$$
$$= \varrho^l c_l^m \left(\frac{x\hat{\imath} - y}{2}\right)^m c^{l-m}$$
$$\times \sum_{\mu=0}^{\lfloor \frac{l-m}{2} \rfloor} \binom{l}{\mu}\binom{l-\mu}{m+\mu}\left(-\frac{x^2+y^2}{4c^2}\right)^\mu \quad (20)$$

where $c = x + y\hat{\imath}$ is the complex variable and $c_l^m$ is normalization factors defined as

$$c_l^m = c_l^{-m} = \frac{\sqrt{(2l+1)(l+m)!(l-m)!}}{l!} \quad (21)$$

while the harmonic polynomials with negative values of $m$ are defined as

$$e_l^{-m}(X) = (-1)^m \overline{e_l^m(X)} \quad (22)$$

The coefficients $q_{kl}^v$ are later determined to guarantee the orthonormality of (19) within the unit ball as

$$q_{kl}^v = \frac{(-1)^k}{2^{2k}} \sqrt{\frac{2l+4k+3}{3}} \binom{2k}{k}(-1)^v \frac{\binom{k}{v}\binom{2(k+l+v)+1}{2k}}{\binom{k+l+v}{k}} \quad (23)$$

The orthogonality relation of 3D Zernike polynomials is defined as

$$\frac{3}{4\pi} \int_{\|X\| \leq 1} Z_{nl}^m(X)\overline{Z_{n'l'}^{m'}(X)}\,dX = \delta_{nn'}\delta_{ll'}\delta^{mm'} \quad (24)$$

### B. Proposed 3D Hahn Moments

Hahn moments were proposed by [63] to solve the problems of continuous orthogonal moments, akin to discrete Chebyshev [64] and weighted Krawtchouk [65] moments. The authors claimed that the resultant Hahn moment has most similar features to the discrete Chebyshev and weighted Krawtchouk moments, although Hahn moments performs better than those moments. This is because discrete Chebyshev and Krawtchouk moments are cases of Hahn moments. A direct implication of this fact is that Hahn moments encompass all the properties of both Chebyshev and Krawtchouk moments, and Hahn moments, in addition, also exhibit intermediate properties between the extremes set by Chebyshev and Krawtchouk moments [66].

For any integer $x \in [0, N-1]$, Hahn polynomial of order $s = 0, 1, \ldots, N-1$ is defined as

$$h_s^{(\mu,\nu)}(a, N) = \sum_{k=0}^{s} \psi_{s,s-k} x^{s-k}$$
$$= (N + \nu - 1)_s(N-1)_s$$
$$\times \sum_{k=0}^{s} (-1)^k \frac{(-s)_k(-a)_k(2N + \mu + \nu - s - 1)_k}{(N + \nu - 1)_k(N-1)_k} \frac{1}{k!} \quad (25)$$

where

$$\psi_{s,k} = \sum_{i=k}^{s} \frac{(-1)^i(-s)_i(\mu + \nu + s + 1)_i}{(\mu + 1)_i(1 - N)_i i!} S_{i-k}^{(k)} \quad (26)$$

with $S_i^{(k)}$ is Stirling numbers of first kind and $(\alpha)_k$ is the Pochhammer symbol defined as

$$(\alpha)_k = \alpha(\alpha + 1)(\alpha + 2) \ldots (\alpha + k - 1) \quad (27)$$

and $\mu > -1, \nu > -1$ are adjustable parameters controlling the shape of polynomials. The discrete Hahn polynomials satisfy the orthogonal condition

$$(\alpha)_k = \alpha(\alpha + 1)(\alpha + 2) \ldots (\alpha + k - 1) \quad (28)$$

where $\rho_s(x)$ is so-called weighting function which is given by

$$\rho_s(a) = \frac{1}{a!(a+\mu)!\,\Gamma(N+\nu-a)\Gamma(N-s-a)} \quad (29)$$

and the square norm $d_s^2$ has the expression

$$d_s^2 = \frac{\Gamma(2N+\mu+\nu-s)}{(2N+\mu+\nu-2s-1)\Gamma(N+\mu+\nu-s)} \times \frac{1}{s!\,\Gamma(N+\mu-s)\Gamma(N+\nu-s)\Gamma(N-s)} \quad (30)$$

To avoid numerical fluctuations in moment computation, the Hahn polynomials are usually scaled by utilizing the square norm and the weighting function, such that

$$\tilde{h}_s^{(\mu,\nu)}(a,N) = h_s^{(\mu,\nu)}(a,N)\sqrt{\frac{\rho_s(a)}{d_s^2}} \quad (31)$$

Therefore, the orthogonality of normalized Hahn polynomials can be described as

$$\sum_{a=0}^{N-1} \tilde{h}_p^{(\mu,\nu)}(a,N)\tilde{h}_q^{(\mu,\nu)}(a,N) = \delta_{pq} \quad (32)$$

Given a digitalized image $f(x,y)$ with size $N \times N$, the $(p+q)$th order of Hahn moment of image is

$$H_{pq} = \sum_{x=0}^{N-1}\sum_{y=0}^{N-1} \tilde{h}_p^{(\mu,\nu)}(x,N)\tilde{h}_q^{(\mu,\nu)}(y,N)f(x,y) \quad (33)$$

Using (25) and (31), the zeroth-order and first-order normalized Hahn polynomials can be easily calculated

$$\tilde{h}_0^{(\mu,\nu)}(a,N) = \sqrt{\frac{\rho_0(a)}{d_0^2}} \quad (34)$$

$$\tilde{h}_1^{(\mu,\nu)}(a,N) = \{(N+\nu-1)(N-1) - (2N+\mu+\nu-2)a\}\sqrt{\frac{\rho_1(a)}{d_1^2}} \quad (35)$$

Higher orders polynomials can be deduced from the following recursive relations,

$$A\tilde{h}_s^{(\mu,\nu)}(a,N) = B\sqrt{\frac{d_{s-1}^2}{d_s^2}}\tilde{h}_{s-1}^{(\mu,\nu)}(a,N) + C\sqrt{\frac{d_{s-2}^2}{d_s^2}}\tilde{h}_{s-2}^{(\mu,\nu)}(a,N) \quad (36)$$

where $s \geq 2$ and

$$A = -\frac{s(2N+\mu+\nu-s)}{(2N+\mu+\nu-2s-1)(2N+\mu+\nu-2s)} \quad (37)$$

$$B = a - \frac{2(N-1)+\nu-\mu}{4} - \frac{(\mu^2-\nu^2)(2N+\mu+\nu)}{4(2N+\mu+\nu-2s+2)(2N+\mu+\nu-2s)} \quad (38)$$

$$C = \frac{(N-s+1)(N-s+\mu+1)}{2N+\mu+\nu-2s+2} \times \frac{(N-s+\nu+1)(N-s+\mu+\nu+1)}{2N+\mu+\nu-2s+1} \quad (39)$$

Eqs. (37)–(39) can be used to efficiently calculate the normalized Hahn moment of any order. The weighting function $\rho_s(a)$ can also be solved by using the recursive relation with respect to $a$

$$\rho_s(a) = \frac{(N-a)(N+\nu-a)}{a(a+\mu)}\rho_s(a-1) \quad a \geq 1 \quad (40)$$

with

$$\rho_s(0) = \frac{1}{\mu!\,\Gamma(N+\nu)\Gamma(N-s)} \quad (41)$$

This study proposes the extension of Hahn moments for 3D images. The proposed 3D Hahn moments is adopting the generalization of $n$-dimensional moments on a cube [45], [46], and defined as

$$H_{pqr} = \sum_{x=0}^{N-1}\sum_{y=0}^{N-1}\sum_{z=0}^{N-1} \tilde{h}_p^{(\mu,\nu)}(x,N)\tilde{h}_q^{(\mu,\nu)}(y,N)\tilde{h}_r^{(\mu,\nu)}(z,N)f(x,y,z) \quad (42)$$

*C. Experimental Setup*

With the goal stated in the section above, an empirical comparative study must be designed and conducted extensively and rigorously. A detailed description of the experimental method is provided in this section.

1) *Dataset Collection*: This section describes the process of transforming molecular structure of ATS drug into 2D and 3D computational data representation, as outlined in [67]. ATS dataset used in this study comes from [1], which contains 60 molecular structures which are commonly distributed for illegal use. On the other hand, 60 non-ATS drug molecular structures, which are randomly collected from [68], is used as benchmarking dataset.

The dataset preparation process is following the procedures outlined by [69]. An example of 2D and 3D molecular structure of ecstasy (3,4-methylenedioxy-methamphetamine), one of notorious ATS drug, and its voxelized molecular structure is shown in Figs. 1, 2, and 3 respectively. After the voxel data has been generated, 3D geometric, complex, Legendre, Zernike, and Hahn moments are calculated up to $8^{th}$ order, which produces 165 features. While the features of 3D geometric, Legendre, and Hahn moments are real numbers, 3D complex and Zernike moments on the other hand are complex numbers.

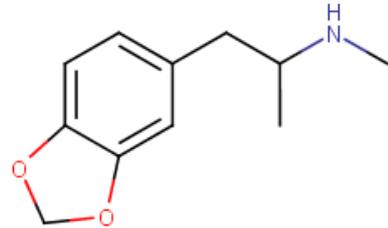

Fig. 1 2D molecular structure of ecstasy drawn using MarvinSketch

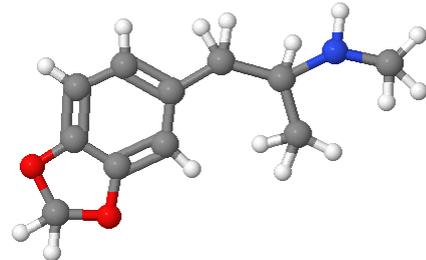

Fig. 2 3D molecular structure of ecstasy converted using Jmol

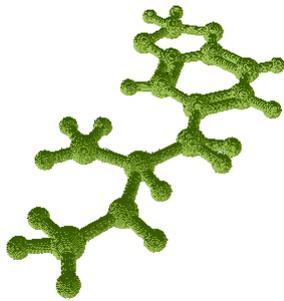

Fig. 3 3D molecular structure of ecstasy voxelized using binvox

TABLE I
CARTESIAN BIT INTERLEAVED VALUES OF ZEROTH-ORDER MOMENTS OF ECSTASY FOR EACH 3D MOMENTS

| 3D Moments | Original Number | Represented Number |
|---|---|---|
| Geometric | 306425 | 42545721700200699567041133799352041472 |
| Complex | 16130711836.218561 | 42576847550484374798153183560267891362 |
| Legendre | 0.000285380519926548 | 14175173924443230618113893434503725056 |
| Zernike | 7708.229987404831 | 42538108148786362155157822007266511528 |
| Hahn | 0.12138471769954105 | 14177782865744079550609449631697511082 |

Therefore, these complex numbers must be transformed into real numbers, because most of pattern recognition tasks only capable to handle real numbers. Ref. [70] proposed a method which consists of four techniques to represent complex number as a real number, and they found Cartesian bit interleaved, one of the proposed technique, as the best representation technique. The value of the zeroth-order moments of ecstasy for each 3D moments represented using Cartesian bit interleaved are shown in Table 1.

2) *Operational Procedure:* The traditional framework of pattern recognition tasks, which are pre-processing, feature extraction, and classification, will be employed in this paper. Therefore, this paper will compare the performance of existing and proposed 3D moments. All extracted instances were tested using training and testing dataset discussed earlier for its processing time, memory consumption, intra- and inter-class variance, and classification of drug molecular structure using leave-one-out classification model, all of which was executed for 50 times.

To justify the quality of features from each moments technique in terms of intra- and inter-class variance, the quartile coefficient of dispersion (QCD) of normalized median absolute deviation (NMAD) is employed. The intra- and inter-class variance is a popular choice of measuring the similarity or dissimilarity of a representation technique [71]–[74]. The QCD measures dispersion and is used to make comparisons within and between data sets [75], and it is defined as

$$QCD_i = \frac{Q3_i - Q1_i}{Q3_i + Q1_i} \qquad (43)$$

where $Q1_i$ and $Q3_i$ are the first and third quartile of the $i$th feature set, respectively. Meanwhile, the median absolute deviation (MAD) is a robust alternative to standard deviation as it is not affected the outliers [76], and it is defined as

$$MAD_i = \text{median}(|X_i - \text{median}(X_i)|) \qquad (44)$$

where $X_i$ is the set of error values attached to the $i$th feature. However, the MAD may be different across different instances, therefore it should be normalized to the original $i$th feature to achieve consistency for different data, such that

$$NMAD_i = \frac{MAD_i}{|x_i|} \times 100\% \qquad (45)$$

In this study, the intra-class variance is defined as the QCD of NMAD for the $i$th feature of a molecular structure compared against intra-class molecular structures, and inter-class variance is defined as the QCD of NMAD for the $i$th feature compared against inter-class molecular structures.

On the other hand, the features are tested in terms of classification accuracy against well-known classifier, Random Forest (RF) [77] from WEKA Machine Learning package [78]. RF is employed in this study, because previous studies conducted by [51], [79]–[81] have found that RF is the most suitable for the molecular structure data. In this study, the number of trees employed by RF is 165, equals to the number of attributes of all 3D moments.

III. RESULTS AND DISCUSSION

The existing and proposed 3D moments will be evaluated numerically in this section to evaluate their merit and quality in representing molecular structure. Table 2 presents the average of processing time, memory consumption, and the intra-class variance ratio relative to the total number of features, while Fig. 4 present the average of classification accuracies from 50 executions.

The results presented in Table 2 and Fig. 4 show that 3D Hahn performs slowest and has lowest value of intra-class variance ratio among other 3D moments, although it requires less memory than Zernike moments and its classification accuracy is higher than 3D complex moments. The slow performance of 3D Hahn is attributed to its polynomial computation which is time consuming, despite using the recursive relationship shown in (36)–(41). Since the classification accuracy is the primary interest of this study, it should also be validated statistically.

TABLE II
AVERAGE OF PROCESSING TIME, MEMORY CONSUMPTION, AND INTRA-CLASS VARIANCE RATIO OF 3D MOMENTS

| 3D Moments | Processing Time (ns/voxel) | Memory Consumption (bytes/ voxel) | Intra-class Variance Ratio |
|---|---|---|---|
| Geometric | 2 | 425 | 92.37% |
| Complex | 39 | 841 | 77.58% |
| Legendre | 16 | 1195 | 67.88% |
| Zernike | 59 | 4405 | 64.24% |
| Hahn | 201 | 1433 | 2.42% |

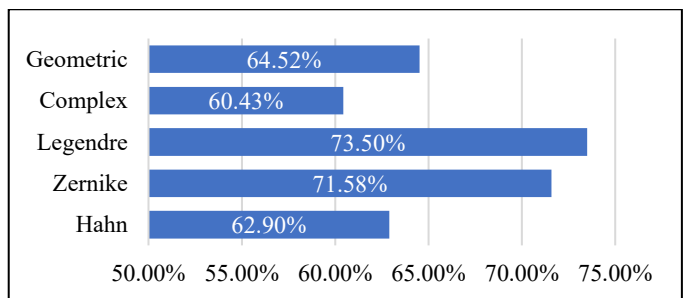

Fig. 4 Average of classification accuracies of 3D moments represented using Cartesian bit interleaved

Prior to performing the statistical validation, the classification accuracy results should be tested for normality. If the results are normally distributed, ANOVA [82] can be used to validate the classification accuracy, otherwise, Kruskal–Wallis $H$ test [83] should be used instead. In this study, the normality of the classification accuracy is tested using Shapiro–Wilk test of normality [84]. The result of the test of normality is presented in Table 3, which shown that the classification accuracy for all 3D moments are normally distributed, since the $p$ value of the Shapiro–Wilk test is greater the 0.05, and thus, ANOVA test can be selected to validate the results. The results of test of homogeneity of variances, ANOVA, and robust tests of equality of means [85] for 3D moments are shown in Tables 4, 5, and 6 respectively.

Based on the results shown in Table 4, there are no homogeneity of variances between groups of 3D moments $(p \leq 0.05)$, therefore the assumption of ANOVA has been violated and the results presented in Table 6 must be considered instead and Games–Howell post-hoc tests [85] to determine the significant difference between pairs must be used consequently. Based on the results presented in Table 6, although both Welch and Brown–Forsythe robust tests of equality of means [85] show that there is statistically significant effect, this study decided to present the Welch robust tests of equality of means.

TABLE III
TESTS OF NORMALITY RESULTS

| 3D Moments | Statistic | $df$ | Sig. ($p$) |
|---|---|---|---|
| Geometric | 0.975 | 50 | 0.376 |
| Complex | 0.978 | 50 | 0.480 |
| Legendre | 0.973 | 50 | 0.305 |
| Zernike | 0.975 | 50 | 0.353 |
| Hahn | 0.960 | 50 | 0.092 |

TABLE IV
TEST OF HOMOGENEITY OF VARIANCES RESULTS

| Levene Statistic | $df$1 | $df$2 | Sig. |
|---|---|---|---|
| 2.918 | 4 | 245 | 0.022 |

TABLE V
ANOVA RESULTS

|  | Sum of Squares | $df$ | Mean Square | $F$ | Sig. |
|---|---|---|---|---|---|
| Between Groups | 0.643 | 4 | 0.161 | 353.376 | 0 |
| Within Groups | 0.111 | 245 | 0 | | |
| Total | 0.754 | 249 | | | |

TABLE VI
ROBUST TESTS OF EQUALITY OF MEANS RESULTS

|  | Statistic[a] | $df$1 | $df$2 | Sig. |
|---|---|---|---|---|
| Welch | 409.479 | 4 | 122.013 | 0 |
| Brown–Forsythe | 353.376 | 4 | 221.078 | 0 |

a. Asymptotically $F$ distributed.

TABLE VII
POST-HOC TEST RESULTS USING GAMES–HOWELL TESTS FOR 3D HAHN MOMENTS VS. OTHER 3D MOMENTS

| Opposing 3D Moments | Mean Difference | Std. Error | Sig. | 95% Confidence Interval | |
|---|---|---|---|---|---|
| | | | | Lower Bound | Upper Bound |
| Geometric | -.01617* | 0.0046 | 0.01 | -0.029 | -0.0033 |
| Complex | .02467* | 0.0039 | 0 | 0.0139 | 0.0354 |
| Legendre | -.10600* | 0.0037 | 0 | -0.1162 | -0.0958 |
| Zernike | -.08683* | 0.0039 | 0 | -0.0978 | -0.0759 |

There is a statistically significant effect of classification accuracy $[F(4, 122.013) = 409.479, p = 0]$ at the $p < 0.05$ level. Post-hoc comparisons using the Games–Howell test shown in Table 7 indicated that the mean score for classification accuracy of 3D Hahn moments $(62.90\% \pm 0.256\%)$ was statistically significantly worse than other candidates $(p < 0.05)$, except with 3D complex moments $(60.43\% \pm 0.289\%, p = 0)$.

Despite providing a not too high performance, this study nevertheless proposes a new 3D moments technique and shows that the proposed 3D Hahn possesses certain potentials to be explored in the future, most notably on its invariance properties.

## IV. CONCLUSION

A new 3D moments technique to represent ATS drug molecular structure has been proposed and the extensive comparative study to the existing 3D moments has been presented in this paper, essentially based on 3D Hahn moments. Despite the experiments have shown that the proposed technique performs rather unexceptionally compared to existing 3D moments in terms of processing time, memory consumption, intra- and inter-class variance, and more importantly, classification accuracy, this study nonetheless serves as a basis towards a better 3D molecular structure representation, especially on using continuous orthogonal moments defined on a cube.

Hence, future works to extend the proposed technique so that it has invariance properties, as well as better representing the molecular structure based on this preliminary study are required. The proposed feature extraction technique will be further validated in the future works using specifically-tailored classifiers for drug shape representation. Furthermore, ATS drug molecular structure data from National Poison Centre, Malaysia, will also be used as additional dataset in the future works.

ACKNOWLEDGMENT

This work was supported by UTeM Postgraduate Fellowship (Zamalah) Scheme and PJP High Impact Research Grant (S01473-PJP/2016/FTMK/HI3) from Universiti Teknikal Malaysia Melaka (UTeM), Malaysia.